\documentclass[pdflatex,sn-mathphys-num]{sn-jnl}

\usepackage{array}
\usepackage{graphicx}%
\usepackage{multirow}%
\usepackage{amsmath,amssymb,amsfonts}%
\usepackage{amsthm}%
\usepackage{mathrsfs}%
\usepackage[title]{appendix}%
\usepackage{xcolor}%
\usepackage{textcomp}%
\usepackage{manyfoot}%
\usepackage{booktabs}%
\usepackage{algorithm}%
\usepackage{algorithmicx}%
\usepackage{algpseudocode}%
\usepackage{listings}%
\usepackage{rotating}
\usepackage{float}

\theoremstyle{thmstyleone}%
\theoremstyle{thmstyletwo}%

\theoremstyle{thmstylethree}%

\begin{document}

\title[Article Title]{PCXR-Copilot: Device-aware pediatric chest radiographic analysis with segmentation, suppression, and edema-related cues}

\author[1]{\fnm{Xueyang} \sur{Li}}
\equalcont{These authors contributed equally to this work.}

\author[2,3]{\fnm{Alec} \sur{Girten}}
\equalcont{These authors contributed equally to this work.}

\author[2]{\fnm{Peng} \sur{Zeng}}

\author[1]{\fnm{Mingze} \sur{Jiang}}

\author[4]{\fnm{Shaoyang} \sur{Huang}}

\author[1]{\fnm{Gelei} \sur{Xu}}


\author[1]{\fnm{Danny Z.} \sur{Chen}}

\author*[1]{\fnm{Yiyu} \sur{Shi}}\email{yshi4@nd.edu}

\author*[2,3]{\fnm{John N.} \sur{Kheir}}\email{john.kheir@childrens.harvard.edu}

\affil[1]{\orgdiv{Department of Computer Science and Engineering}, \orgname{University of Notre Dame}, \orgaddress{\city{Notre Dame}, \state{IN}, \country{USA}}}

\affil[2]{\orgdiv{Department of Cardiology}, \orgname{Boston Children's Hospital}, \orgaddress{\city{Boston}, \state{MA}, \country{USA}}}

\affil[3]{\orgdiv{Department of Pediatrics}, \orgname{Harvard Medical School}, \orgaddress{\city{Boston}, \state{MA}, \country{USA}}}

\affil[4]{\orgdiv{School of Computing and Information}, \orgname{University of Pittsburgh}, \orgaddress{\city{Pittsburgh}, \state{PA}, \country{USA}}}
\abstract{Patients admitted to pediatric ICU and cardiac ICU often require intensive perioperative support and frequent chest radiography to monitor support devices, extravascular lung water, and body-wall edema. Radiographic interpretation and analysis is often difficult due to superimposed anatomic and iatrogenic structures that obscure clinically relevant signals. While deep learning can provide decision support, translation to pediatric radiographs is still limited by high device burden and scarcity of pediatric datasets with clinically grounded device annotations. To address this gap, we first introduce \textbf{PediaDeviceSeg-1K}, a curated dataset of 1,000 anterior-posterior radiographs of pediatric CICU patients with clinically grounded, fine-grained annotations of common ICU devices. To our best knowledge, this is the first publicly available pediatric CICU CXR resource with clinically grounded device annotations for device-focused benchmarking. Building on \textbf{PediaDeviceSeg-1K}, we further present \textbf{PCXR-Copilot}, a device-aware pipeline that outputs lung and device masks, optional device-suppressed views, and edema-related visualization maps and scalar indices. The full pipeline has been integrated and runs in real time within the TWIST platform at Boston Children’s Hospital, demonstrating feasibility for real-world clinical deployment.

}

\keywords{Pediatric chest radiograph, pediatric ICU, clinical decision support, medical device segmentation}

\maketitle

\section{Introduction}
The increasing prevalence of children with congenital and acquired cardiac conditions, combined with advances in surgical and medical management of these conditions, has led to improved outcomes but also increased medical complexity~\cite{rossano2012prevalence, oster2013temporal, vey2025current, amdani2022burden, kheiwa2025worldwide}. With this increased complexity, the field has seen an increase in the pre-, peri-, and post-surgical use of support devices, such as endotracheal tubes (ETT), cardiac pacing wires, surgical drains, implanted cardiac devices, chest tubes, enteric tubes, and central venous catheters, as part of the routine care for infants and children admitted to Pediatric Intensive Care Units (PICUs) and Cardiac 
Intensive Care Units (CICUs). Day-to-day care of these patients regularly involves routine -- often daily -- chest radiography to assess the cardiac silhouette, degree of extravascular lung water (EVLW) or pulmonary edema, degree of body wall edema, and positions of non-biologic support devices. Specifically, quantifying the amount of EVLW has important prognostic implications on the clinical status of patients with heart disease in various stages of surgical or medical management~\cite{kaskinen2017assessment, sampaio2015influence}.  

Unfortunately, interpretive errors are a well established phenomenon in the analysis of radiographic studies including chest radiography~\cite{brady2017error}. Such errors have been attributed as the result of both cognitive (bias, heuristics, premature closure, etc.) and systemic (lighting, fatigue, repetitive tasks, etc.) factors involved in the interpretation of radiographic studies~\cite{lee2013cognitive}. In addition, added visual information on chest radiographs of pediatric cardiac surgical patients, including both biologic structures (i.e., ribs) and support devices, obscures and increases the difficulty in assessing EVLW and body wall edema. Standard chest radiography interpretation has also shown mixed to limited accuracy in quantifying EVLW~\cite{schulz2023deep, halperin1985evaluation}.  

Deep learning (DL) methods offer a promising avenue to mitigate these interpretive challenges by extracting structural information from chest radiographs and highlighting clinically relevant patterns. Prior work has demonstrated the utility of chest X-ray (CXR)-based models for assessing pulmonary edema and related surrogates of extravascular lung water, largely in adult cohorts~\cite{schulz2023deep, koul2023artificial, park2024extraction, horng2021deep}. Complementary efforts have addressed detection and segmentation of indwelling support devices and postoperative hardware~\cite{tang2023analysis, an2024artificial, eibschutz2024artificial, wongveerasin2024deep}. However, pediatric ICU radiographs present distinctive barriers to translation, including high device burden, substantial occlusion by non-biologic hardware, and, most importantly, limited availability of pediatric datasets with fine-grained device annotations for reproducible development and benchmarking.

To address these gaps, we first present \textbf{PediaDeviceSeg-1K}, which to the best of our knowledge is the first publicly released pediatric CXR dataset with fine-grained, clinically grounded annotations of common ICU devices, comprising 1{,}000 frontal radiographs obtained from routine pediatric CICU practice. We will release this resource to enable reproducible benchmarking, confounder-aware modeling, and other downstream research. Building on PediaDeviceSeg-1K, we present \textbf{PCXR-Copilot}, a device-aware pediatric CXR assistance pipeline that produces structured, inspectable outputs for decision support, including lung masks, medical device masks, optional device-suppressed radiographs, and edema-related cues comprising visualization maps and scalar indices for pulmonary and body-wall edema. \textbf{Together, these contributions establish a transparent, device-aware workflow that supports clinician inspection and longitudinal reasoning in high-complexity pediatric radiographs, while providing a benchmark dataset to accelerate and standardize future method development.}


\section{Results}
In this section, we present results for our two primary contributions: (1) a curated pediatric chest radiograph dataset with expert device annotations (PediaDeviceSeg-1K), and (2) an end-to-end assistance pipeline (PCXR-Copilot) integrated into a clinical information processing and visualization platform at Boston Children’s Hospital called TWIST. This is a provider-facing platform that manifests multimodal data including real-time-transformed data from electronic health records, bedside monitoring and medical devices (e.g., ventilators, ECMO pumps), and radiographic imaging. We first summarize the cohort composition, annotation protocol, and device prevalence in the 1,000-image dataset. We then report quantitative performance of the core perception modules, including medical device segmentation and lung segmentation, and evaluate device suppression with an emphasis on removal effectiveness and structural preservation. Finally, we characterize the pipeline’s edema-related outputs, including pulmonary edema and body-wall edema highlighting and a quantitative pulmonary edema index, and examine the feasibility of real-world deployment in TWIST based on runtime and integration characteristics. Through the analyses, we emphasize robustness across common device types and imaging conditions, and we document representative successes and failure modes to inform safe clinical translation.


\subsection{PediaDeviceSeg-1K dataset summary}

The source cohort comprised 105 patients with a total of 3,122 frontal pediatric chest radiographs. From these radiographs, we selected 1,000 images for manual device annotation. To improve the representativeness and efficiency of the curation process, we combined automated representative selection with targeted expert review. Specifically, images were first prioritized using a representative selection strategy that aims to reduce redundancy by favoring samples that are both informative and diverse, following the principles of representative sample selection in prior work~\cite{zheng2020annotation}.
Candidate images were then screened by human reviewers to confirm the presence of at least one target support device, with the annotation focusing on clinically relevant, device-instrumented radiographs. After label harmonization (mapping and casing unification), per-image label appearance counts were obtained and summarized in Table~\ref{tab:device_summary}. The resulting dataset was highly device-dense and heterogeneous: each radiograph contains a median of 19 device instances (IQR, 11--29) spanning 6 distinct device categories (IQR, 4--9), and 90.4\% of the images contained at least two device categories. Additional descriptors of device burden and device--lung spatial overlap are reported in Table~\ref{tab:dataset_heterogeneity}.

\begin{table*}[t]
    \centering
    \caption{\textbf{Medical device label prevalence in the annotated pediatric CXR dataset.} The counts indicate the number of labeled images (out of 1{,}000) in which each label appears at least once (unique per image) after label mapping and casing unification.}
    \label{tab:device_summary}
    \small
    \setlength{\tabcolsep}{4pt}
    \begin{tabular}{@{}>{\raggedright\arraybackslash}p{5.35cm} r >{\raggedright\arraybackslash}p{5.35cm} r@{}}
    \toprule
    \textbf{Label} & \textbf{n} & \textbf{Label} & \textbf{n} \\
    \midrule
    Sternal Wire & 732 & End tidal CO2 monitor & 72 \\
    Nasogastric Tube & 578 & Vent Tubing & 61 \\
    EKG Lead & 536 & Tourniquet clip & 56 \\
    EKG sticker & 489 & Left internal jugular vein catheter & 48 \\
    Temporary pacing wires & 427 & Esophageal temperature probe & 41 \\
    Endotracheal tube (ETT) & 423 & Cardiac tourniquet & 35 \\
    Chest Tube & 354 & Pilot balloon (on ETT) & 21 \\
    Vascular clip & 313 & Wound vac tubing & 21 \\
    Miscellaneous cable & 285 & Central venous catheter (other site) & 18 \\
    Intracardiac line & 276 & Pilot balloon tubing & 13 \\
    Coil & 254 & Atrial septal defect closure device & 12 \\
    Right internal jugular vein catheter & 221 & Prosthetic valve & 2 \\
    Stent & 183 & Umbilical venous catheter (UVC) & 2 \\
    Permanent pacemaker (PPM) Wire & 173 & G tube button & 1 \\
    Carina & 160 & G tube tubing & 1 \\
    Permanent pacemaker (PPM) generator & 136 & Pigtail chest drain & 1 \\
    Peripherally inserted central catheter (PICC) & 100 & Wound vac sponge & 1 \\
    \bottomrule
    \end{tabular}
\end{table*}

\begin{table*}[t]
    \centering
    \caption{\textbf{Dataset complexity and device-burden descriptors.} Summary statistics are computed from the labeled set (1,000 images). When applicable, values are reported as median [IQR] or mean $\pm$ SD. Device pixel ratio is the percentage of image pixels occupied by devices; device-in-lung ratio is the percentage of image pixels occupied by devices within the lung masks; device-lung overlap is the percentage of lung pixels occupied by devices.}
    \label{tab:dataset_heterogeneity}
    \small
    \setlength{\tabcolsep}{4pt}
    \begin{tabular}{@{}>{\raggedright\arraybackslash}p{5.0cm} >{\centering\arraybackslash}p{2.8cm} >{\centering\arraybackslash}p{4.0cm}@{}}
    \toprule
    \textbf{Descriptor} & \textbf{Overall} & \textbf{Across folds (mean $\pm$ SD)} \\
    \midrule
    Patients, $n$ & 100 & 20.0 $\pm$ 0.0 \\
    Images, $n$ & 1000 & 200 $\pm$ 0.9 \\
    Images per patient & 6.0 [3.0--14.0] & 5.8 $\pm$ 0.3 \\
    Device categories per image & 6.0 [4.0--9.0] & 6.4 $\pm$ 0.5 \\
    Device instances per image & 19.0 [11.0--29.0] & 18.8 $\pm$ 1.3 \\
    Device pixel ratio (\%) & 2.70 [1.34--4.51] & 2.67 $\pm$ 0.28 \\
    Device-in-lung ratio (\%) & 0.23 [0.09--0.43] & 0.24 $\pm$ 0.04 \\
    Device--lung overlap (\%) & 2.23 [0.78--4.20] & 2.28 $\pm$ 0.51 \\
    Heavy-device cases (device pixel ratio $> 5\%$), \% & 19.4 & 19.4 $\pm$ 4.2 \\
    Multi-device cases ($\geq 2$ device categories), \% & 90.4 & 90.4 $\pm$ 2.2 \\
    \bottomrule
    \end{tabular}
\end{table*}

\begin{table*}[t]
\centering
\caption{\textbf{Medical device segmentation results.}
Binary segmentation is reported as Dice. Multi-label segmentation is evaluated on device categories with prevalence $\geq 30\%$; Dice is macro-averaged over included labels. Values are reported as mean $\pm$ SD across five folds.}
\label{tab:devseg_models}
\resizebox{\textwidth}{!}{%
\begin{tabular}{l l c}
\toprule
\textbf{Task} & \textbf{Model} & \textbf{Dice (\%)}\\
\midrule
\multirow{5}{*}{Binary (device vs.\ background)}
                                                & nnU-Net & $77.8 \pm 1.8$\\
                                                & GLCP & $77.5 \pm 1.8$\\
                                                & U-Net++ (EfficientNet-b0 encoded) & $76.2 \pm 1.6$\\
                                                & U-Net++ (ResNet-18 encoded) & $75.7 \pm 1.8$\\
                                                & DeepLabv3+ & $74.8 \pm 1.4$\\
\midrule
\multirow{5}{*}{Multi-label (prevalence $\geq 30\%$)}
                                                & nnU-Net & $68.2 \pm 2.7$\\
                                                & GLCP & $65.1 \pm 2.0$\\
                                                & U-Net++ (ResNet-18 encoded) & $55.9 \pm 1.2$\\
                                                & U-Net++ (EfficientNet-b0 encoded) & $58.4 \pm 2.6$\\
                                                & DeepLabv3+ & $61.1 \pm 2.1$\\
\bottomrule
\end{tabular}
}
\end{table*}

\begin{figure*}[t]
    \centering
    \includegraphics[width=0.9\textwidth]{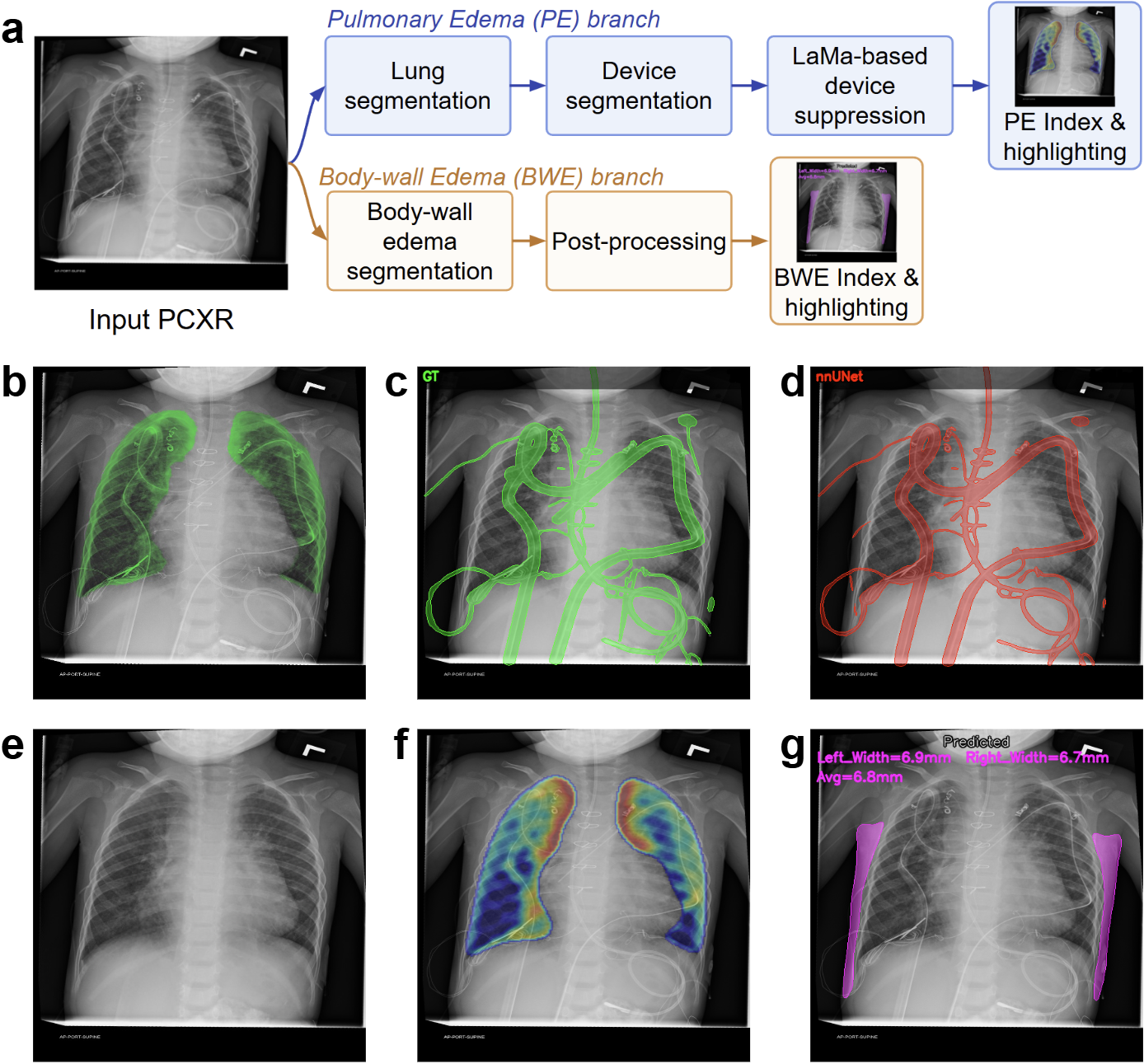}
    \caption{\textbf{PCXR-Copilot end-to-end pipeline and representative outputs.}
        (a) Overview of the PCXR-Copilot pipeline, showing the example input pediatric chest radiograph (PCXR) and the two analysis branches for pulmonary edema (PE) and body-wall edema (BWE).
        (b) Predicted lung masks overlaid on the input radiograph.
        (c) Expert device annotations (ground truth) for the input radiograph.
        (d) Device masks predicted by nnU-Net.
        (e) Device-suppressed image produced by mask-guided inpainting.
        (f) Pulmonary edema highlighting with the derived PE index.
        (g) Body-wall edema highlighting with the derived width-based index.}
    \label{fig:pipeline_overview}
\end{figure*}


\begin{figure*}[t]
    \centering
    \includegraphics[width=\textwidth]{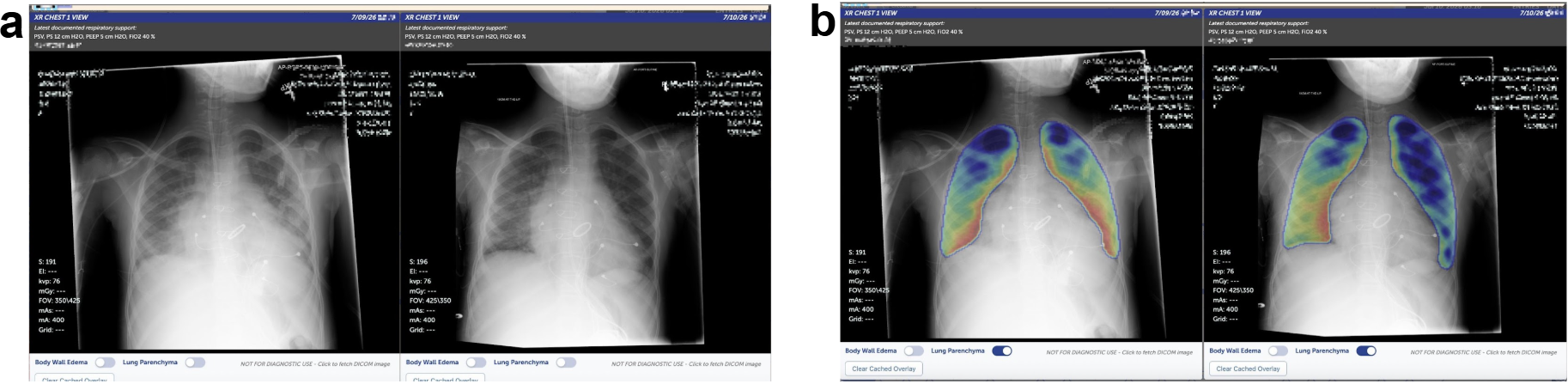}
    \caption{\textbf{Lung water overlay in the TWIST radiology viewer.}
    Screenshots of the TWIST radiology viewer show comparative chest radiographs
    of the same patient separated by 28 hours, in chronological order from left to right.
    (a) Two radiographs displayed without overlay.
    (b) The same two radiographs with an overlay depicting lung water, which
    facilitates recognition of the decrease in lung water between the two radiographs.
    Note the handling of the PICC line in the right upper extremity and the pilot
    balloon, both of which are subtracted from the lung water overlay computation.}
    \label{fig:twist_overlay}
\end{figure*}

\begin{figure*}[t] 
    \centering 
    \includegraphics[width=0.8\linewidth]{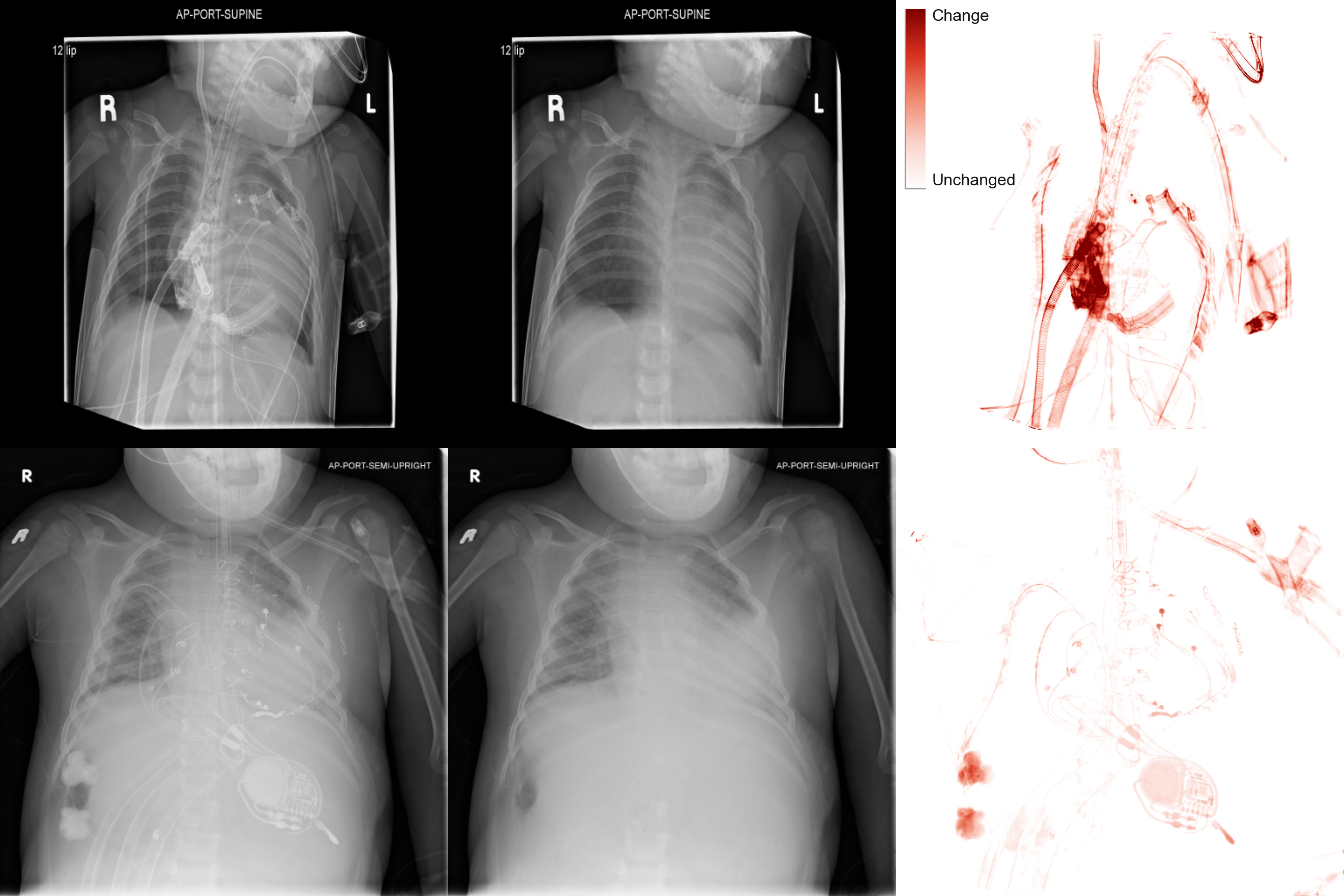} 
    \caption{\textbf{Representative medical device suppression and associated image changes.} For each case, the original radiograph (left), device-suppressed radiograph (middle), and pixel-wise difference map (right) are shown. Difference maps highlight regions altered by suppression, with changes concentrated predominantly along device-containing regions and limited changes in surrounding non-device anatomy.} 
    \label{fig:suppression_example} 
\end{figure*}

\begin{figure*}[t]
    \centering
    \includegraphics[width=\textwidth]{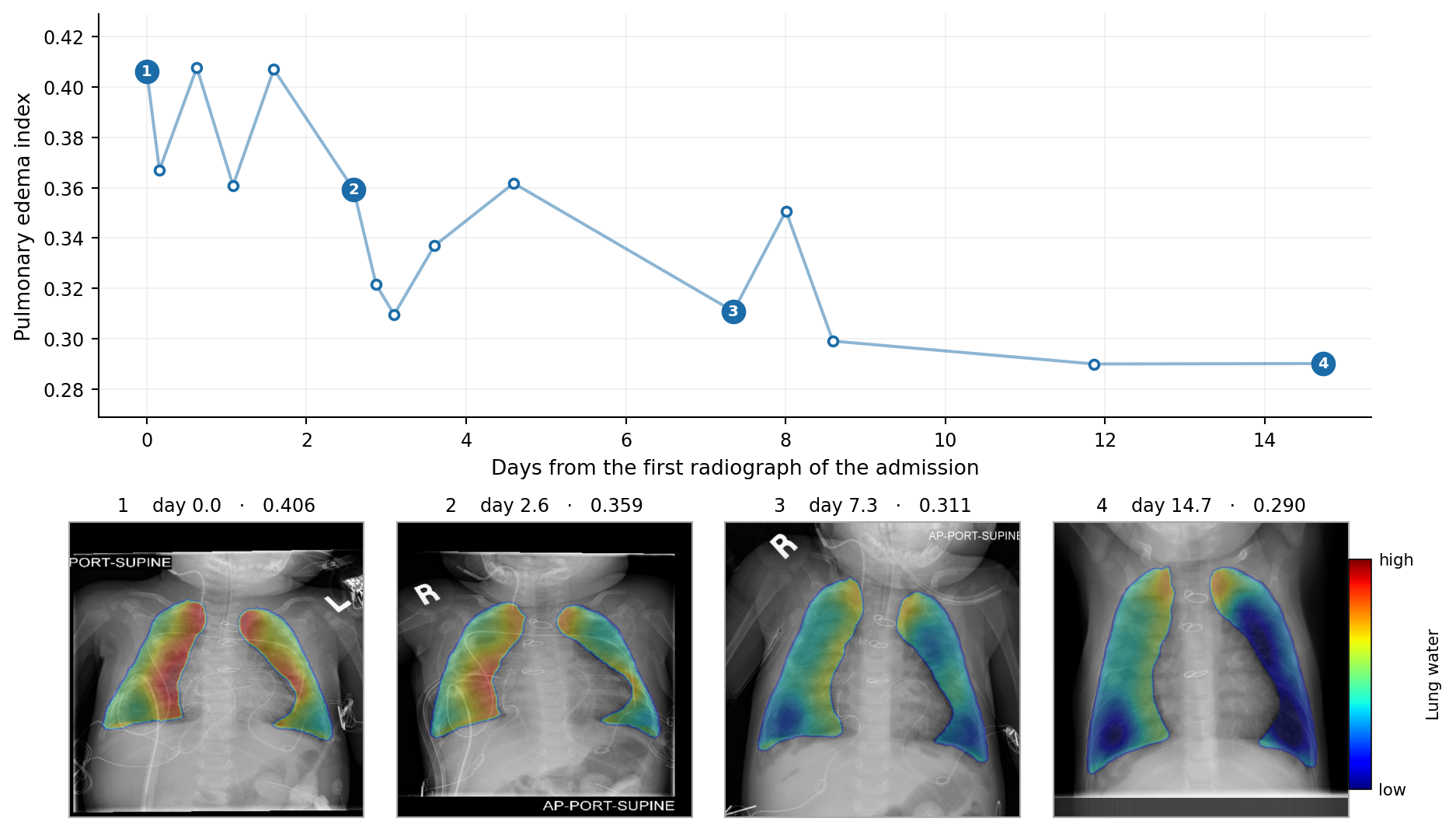}
    \caption{\textbf{Within-patient evolution of the pulmonary edema index.}
    Index values for 15 frontal radiographs acquired during a single admission;
    numbered markers denote the four timestamps shown below. Lung-water overlays are
    restricted to the predicted lung fields and share a common color scale across the
    panels.}
    \label{fig:pe_case}
\end{figure*}



\subsection{PCXR-Copilot pipeline}
In this subsection, we report end-to-end results of our PCXR-Copilot, emphasizing how the individual modules operate together to produce clinically interpretable outputs from a single frontal pediatric CXR. Fig.~\ref{fig:pipeline_overview} summarizes the pipeline and illustrates representative outputs, including lung and device segmentation, device suppression, pulmonary edema cue evidence, body-wall edema highlighting with a quantitative index, and deployment within the TWIST clinical platform. The following subsubsections quantify module performance where reference annotations are available and provide qualitative examples to 
illustrate typical behavior and failure modes in real-world, device-instrumented radiographs.

\subsubsection{Lung segmentation}

The lung segmentation module delineates the pulmonary fields used for downstream edema-related visualization and quantification. Evaluated using five-fold cross-validation on 300 manually annotated pediatric CXRs, U-Net++~\cite{zhou2018unet++} achieved a Dice score of $92.7 \pm 2.7\%$ (mean $\pm$ SD across folds). The resulting lung masks were used to define the regions of interest for pulmonary edema highlighting, PE-index computation, and device--lung overlap analysis.

\subsubsection{Medical device segmentation and suppression}

\noindent \textbf{Device Segmentation.} We formulated medical device segmentation primarily as a binary task (device vs.\ background) to generate a union-of-devices mask for downstream suppression and edema-related analyses, where the presence of any device acts as an occluder and intensity confounder. Evaluated using five-fold cross-validation, nnU-Net achieved the highest mean Dice for binary device segmentation at $77.8 \pm 1.8\%$, closely followed by GLCP~\cite{zhou2025glcp} at $77.5 \pm 1.8\%$ (Table~\ref{tab:devseg_models}). U-Net++~\cite{zhou2018unet++} with EfficientNet-b0~\cite{han2021rethinking} and ResNet-18~\cite{he2016deep} encoders achieved $76.2 \pm 1.6\%$ and $75.7 \pm 1.8\%$, respectively, while DeepLabv3+~\cite{chen2018encoder} achieved $74.8 \pm 1.4\%$. Based on its highest mean binary Dice and direct suitability for generating category-agnostic masks required by downstream processing, nnU-Net was taken as the default binary segmentation model in PCXR-Copilot.

We also explored multi-label device segmentation to predict device categories explicitly. However, device prevalence in our cohort followed a pronounced long-tail distribution, as summarized in Table~\ref{tab:device_summary}, limiting reliable learning for rare classes. To ensure sufficient supervision per label, we restricted multi-label evaluation to device categories appearing in at least 30\% of images. Under this setting, nnU-Net achieved the highest macro-averaged Dice at $68.2 \pm 2.7\%$, followed by GLCP at $65.1 \pm 2.0\%$, as reported in Table~\ref{tab:devseg_models}. Given that downstream device suppression and edema-related processing require only a category-agnostic union-of-devices mask, we used the binary nnU-Net model for the remainder of the pipeline analysis. We anticipate that category-specific segmentation will benefit from larger per-class sample sizes and will be revisited as additional labeled cases become available.

\noindent \textbf{Device Suppression.} We then assessed medical device removal using the predicted device masks to help the mask-guided inpainting module. Because pixel-aligned, device-free reference images are not available in routine clinical acquisition, suppression quality cannot be evaluated directly against a ground-truth target. We therefore assessed both the reduction of device-associated signal within the reconstructed regions and the preservation of image content outside these regions. Across the 1,000 annotated radiographs for which both the annotated device regions and corresponding dilated inpainting masks were available, mean intensity within the annotated device regions decreased from 151.1 to 132.0 on the 0--255 grayscale range, corresponding to a 12.5\% reduction. The mean intensity difference between device regions and surrounding unaffected tissue regions decreased from 29.3 to 9.6 gray levels, representing a 67\% reduction in local device-to-tissue contrast. In comparison, the mean absolute intensity difference outside the dilated inpainting masks was only 0.83 gray levels (0.3\% of the grayscale range). This nonzero residual does not reflect modification by the inpainting module, but was actually incurred from the image-comparison procedure: the native radiographs were up to approximately 2,450 pixels wide, whereas the inpainting network operated at $512\times512$ resolution, requiring the original and reconstructed images to be resampled into a common grid, thereby introducing small resampling and compression artifacts. 
Outside the reconstructed regions, the original image content was therefore unaltered by the inpainting operation.

These quantitative findings are also reflected visually in Fig.~\ref{fig:suppression_example}: pixel-wise difference maps show that changes are concentrated primarily along device-containing regions, whereas most surrounding non-device anatomy remains largely unchanged. Residual device-associated signal and local reconstruction artifacts were observed in some cases, particularly when device masks under-segmented thin structures, when multiple devices overlapped within the lung fields, or when strong edges near the mediastinum created ambiguous texture continuation for inpainting. To support safe use and downstream quality control, both the predicted device masks and the corresponding suppressed images are made available for user inspection.

\subsubsection{Pulmonary edema quantification and highlighting}
The pipeline produces two edema-related outputs within the lung fields: (i) a Grad-CAM based pulmonary edema (PE) highlight map that provides an interpretable visualization of image regions 
associated with the model’s PE-related signal, and (ii) a scalar PE index that summarizes this signal within the predicted lung masks. Representative outputs as 
seen in the TWIST viewer, including the within-patient change over 28 hours, are shown in Fig.~\ref{fig:twist_overlay}. Because pixel-level quantitative reference standards for edema extent are not routinely acquired in clinical practice, here we focus on characterizing the behavior, dynamic range, and robustness of these outputs rather than claiming quantitative diagnostic accuracy.



We first characterized cohort-level variability of the pulmonary edema index to establish its dynamic range. In our pediatric cohort, the normalized index had a median of 0.358 (interquartile range, 0.337--0.381) and a mean of 0.358 $\pm$ 0.038. We summarize dispersion using the median and interquartile range (IQR) alongside the mean and standard deviation, and the observed spread indicates a non-trivial dynamic range that may support longitudinal tracking and cohort-level stratification. We further quantified the impact of suppression and normalization using paired, image-level comparisons. Relative to the original index, suppressing ribs and medical devices reduced the index, with a median paired change of $-0.0629$ and an IQR from $-0.0715$ to $-0.0539$, consistent with removing high-intensity non-parenchymal contributions. Technique normalization subsequently increased the rib- and device-excluded index, with a median paired change of $+0.0146$ and an IQR from $+0.0091$ to $+0.0212$, improving comparability across acquisition conditions. Together, suppression shifts the index downward while normalization re-centers values, which is expected to reduce technique-driven variability. The index also varies within a single admission: in the example shown in Fig.~\ref{fig:pe_case}, it decreased from 0.406 to 0.290 across 15 radiographs spanning 14.7 days. This downward trend is visually corroborated by the lung-water overlays, which show a progressive reduction in signal intensity across the four timestamps, consistent with the decrease in the pulmonary edema index. To provide an exploratory clinical reference for relative within-patient change, we conducted a paired assessment by a pediatric cardiac intensivist who was blinded to the PE index. Among the 66 radiograph pairs receiving a directional judgment, the signed PE-index difference discriminated the reader-assigned direction of greater pulmonary edema with an AUC of 0.806.

\begin{figure*}[t] \centering 
    \includegraphics[width=0.85\textwidth]{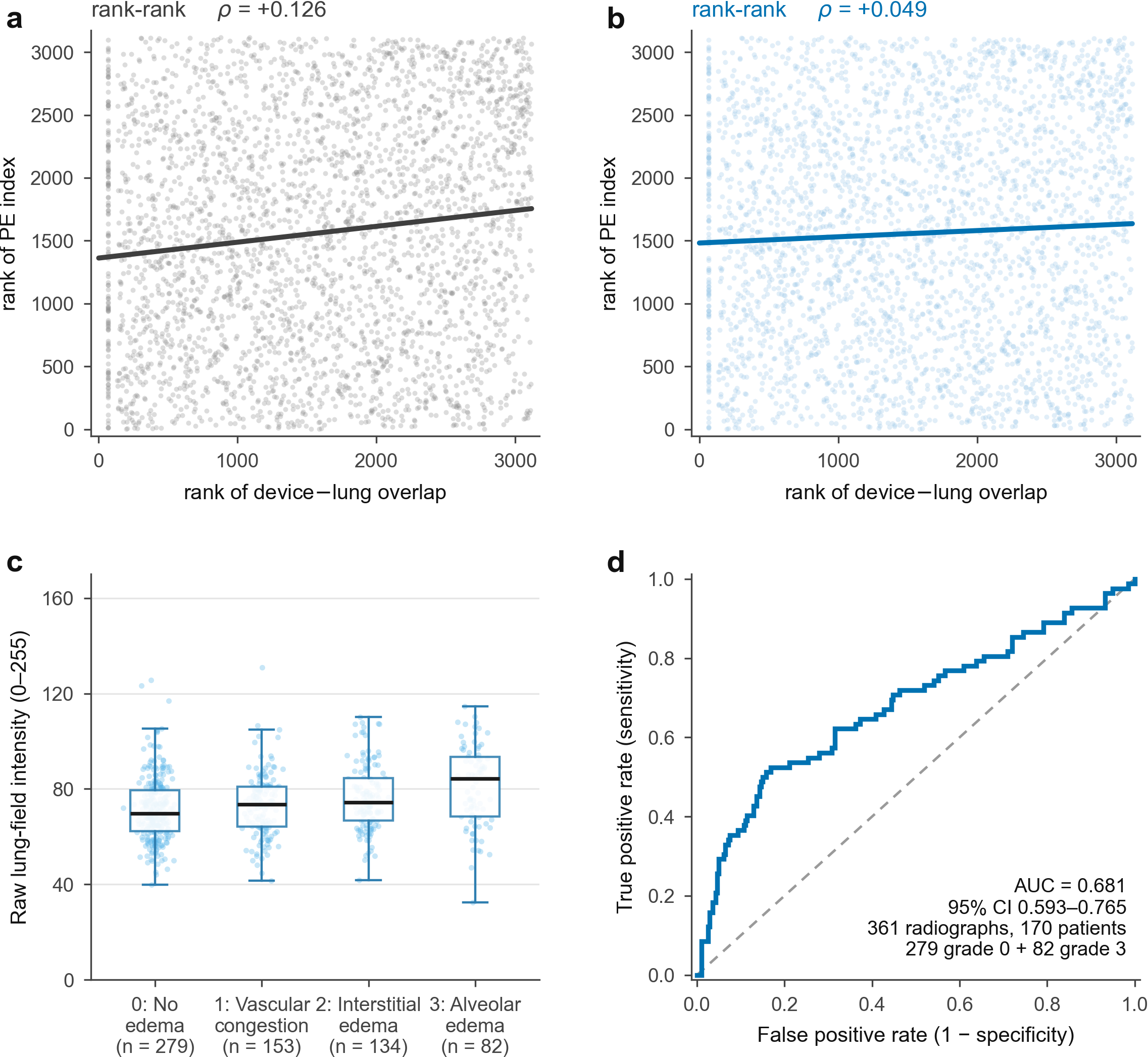} 
    \caption{\textbf{Device-related dependence and external evaluation of the PE index.}
    (\textbf{a,b}) Rank--rank plots show the association between device--lung overlap and the PE index in the BCH cohort with devices retained (a) and suppressed (b). Solid lines show least-squares fits to the ranked values; Spearman $\rho$ decreased from 0.126 to 0.049 after suppression.
    (\textbf{c}) Distribution of the lung-field intensity index across edema grades in MIMIC-CXR, using the BCH-trained lung segmentation model: grade 0, no edema ($n=279$); grade 1, vascular congestion ($n=153$); grade 2, interstitial edema ($n=134$); and grade 3, alveolar edema ($n=82$). Boxes show medians and interquartile ranges; whiskers extend to the most extreme observations within 1.5 interquartile ranges, and points represent individual radiographs.
    (\textbf{d}) ROC curve for grade 3 versus grade 0, based on 361 radiographs from 170 patients. The 95\% confidence intervals for Spearman $\rho$ and AUC were obtained from 2,000 patient-level bootstrap resamples.}
    \label{fig:pe_device_corr} 
\end{figure*}

To evaluate whether device suppression reduces the sensitivity of the PE index to intrapulmonary device burden, we compared the association between device--lung overlap and the PE index with devices either retained or suppressed (Fig.~\ref{fig:pe_device_corr}a,b). Device--lung overlap was defined as the fraction of predicted lung pixels occupied by predicted device pixels. The same radiographs and all other processing steps were used in both conditions, with only device suppression toggled. With devices retained, the PE index showed a weak positive association with device--lung overlap (Spearman $\rho=0.126$); after suppression, this association decreased to $\rho=0.049$ ($\Delta\rho=-0.077$), indicating reduced device-related dependence of the PE index.

Given the scarcity of publicly available pediatric chest radiographs with pulmonary edema severity annotations, we assessed cross-population transfer using an external adult MIMIC-CXR cohort~\cite{johnson2019mimic,liao2021edemagrades} comprising 648 eligible radiographs from 268 patients. Our existing pediatric lung segmentation model was applied without retraining or fine-tuning, and the intensity index was computed without supervised fitting to edema labels. Despite the pediatric-to-adult domain shift, the index showed a positive association with four-level pulmonary edema severity (Spearman $\rho=0.211$, 95\% CI 0.120--0.301; Fig.~\ref{fig:pe_device_corr}c). For discrimination between grade 0 (no edema; 279 radiographs) and grade 3 (alveolar edema; 82 radiographs), the AUC was 0.681 (95\% CI 0.593--0.765; Fig.~\ref{fig:pe_device_corr}d). These findings demonstrate that an edema-related intensity signal remained detectable across populations without additional model training on the external cohort.

\begin{figure}[t]
  \centering
  \includegraphics[width=0.8\linewidth]{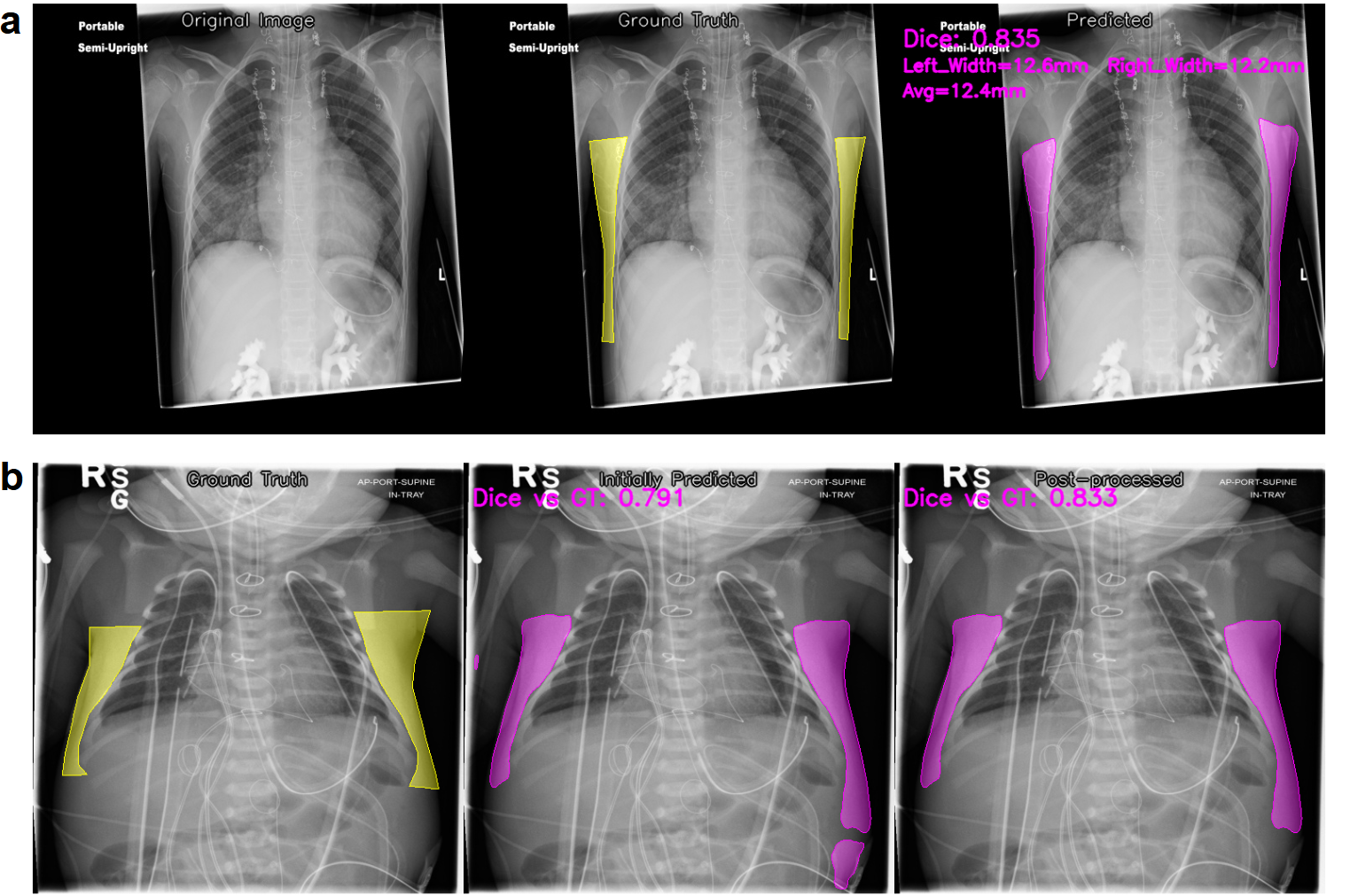}
  \caption{Qualitative visualization of body-wall edema region segmentation and post-processing.
  (a) Representative chest radiographs showing the original image, manual ground-truth annotation (yellow overlay), and model prediction (magenta overlay). The predicted panel reports the Dice score against ground truth and 
  the left, right, and average body-wall width estimates (in mm).
  (b) A post-processing comparison example, showing ground truth (yellow), the initial prediction before post-processing, and the post-processed prediction after retaining the two longest connected components and removing small spurious regions. The post-processing reduces isolated false positives and improves overlap with ground truth, as reflected by the Dice score.}
  \label{fig:bwe}
\end{figure}
\begin{figure}[t]
    \centering
    \includegraphics[width=\linewidth]{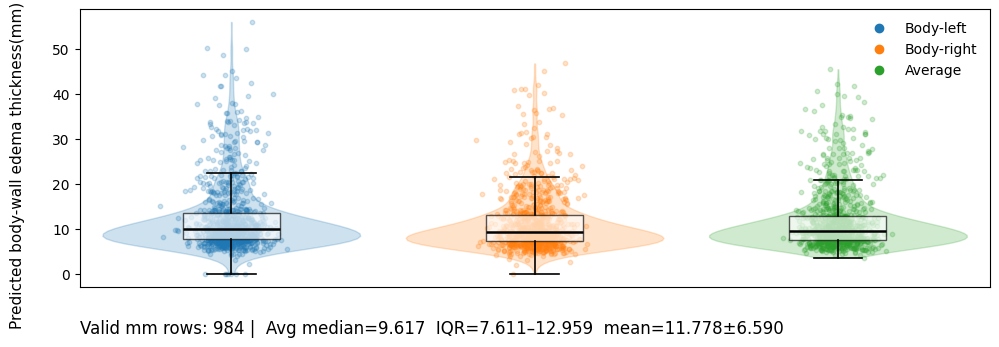}
    \caption{\textbf{Cohort-level distribution of predicted body-wall edema thickness.}
    Violin and box plots summarize the radiograph-level distribution of predicted body-wall edema thickness (in mm) for the body-left and body-right regions and their average. Points are horizontally jittered within each category for visualization.}
    \label{fig:bwe_distribution}
\end{figure}

\subsubsection{Body-wall edema highlighting}
We additionally generate body-wall edema highlight maps using a supervised U-Net++ segmentation model. In five-fold cross-validation, the model achieved a Dice score of 78.8 $\pm$ 2.3. Fig.~\ref{fig:bwe}a shows representative outputs that localize predicted body-wall edema regions on frontal pediatric chest radiographs. In addition, Fig.~\ref{fig:bwe}b illustrates the post-processing strategy used to enforce a clinically consistent bilateral output, which yielded cleaner segmentation and improved overlap with reference annotations. To characterize cohort-level prevalence and variability of the predicted body-wall edema signal, we summarize the radiograph-level distribution of the estimated body-wall edema thickness in Fig.~\ref{fig:bwe_distribution}. For the average thickness, the median was 9.62\,mm with an IQR of 7.61 to 13.00\,mm, and the mean was 11.78$\pm$6.60\,mm. Because quantitative, pixel-level reference annotations for body-wall edema are not routinely acquired in clinical practice, we report these results as descriptive outputs intended to support visualization and clinical decisions, rather than as a definitive quantitative endpoint.



\subsubsection{Clinical deployment in TWIST}
To support prospective clinical translation, the PCXR-Copilot pipeline has been integrated into TWIST, enabling clinicians to view lung water heatmaps and quantified body-wall edema alongside other relevant clinical information, such as ventilator settings, fluid balance, and medications. Fig.~\ref{fig:twist_overlay} shows representative TWIST interface screenshots.


\section{Discussion}
\label{sec:discussion}


In this study, we present PCXR-Copilot, a device-aware pediatric chest radiograph assistance pipeline that produces structured, inspectable outputs for decision support, including lung masks, medical device masks, optional device-suppressed radiographs, and edema-related cues comprising visualization maps and scalar indices for pulmonary and body-wall edema. We additionally introduce PediaDeviceSeg-1K, a curated dataset of 1,000 frontal pediatric CXRs with detailed device annotations derived from real-world CICU practice, which we will make publicly available to support reproducible benchmarking and downstream research.

These contributions address a clinically important need: children with congenital and acquired cardiac conditions admitted to PICU and CICU often require intensive perioperative monitoring and substantial non-biologic device support, resulting in radiographs with high visual complexity~\cite{rossano2012prevalence, oster2013temporal, vey2025current, amdani2022burden, kheiwa2025worldwide}. Routine, frequently daily, imaging is used to track cardiopulmonary status, including EVLW (pulmonary edema) and body-wall edema, alongside verification of device positioning, yet interpretation can be challenging due to both cognitive and system-level factors and occluding structures such as ribs and indwelling hardware~\cite{lee2013cognitive}. Given the prognostic relevance of edema-related findings and the practical difficulty of quantitative assessment in standard clinical workflows~\cite{schulz2023deep, halperin1985evaluation}, PCXR-Copilot is designed to provide complementary quantitative indicators and visual evidence, while retaining transparency through explicit masks and intermediate outputs that facilitate clinician inspection and quality control.

In addition to the pipeline, the dataset PediaDeviceSeg-1K addresses a key data gap in pediatric imaging. Compared with adult chest radiography resources, publicly available pediatric CXR datasets remain relatively limited, which can hinder the development and rigorous evaluation of deep learning methods in this domain. Moreover, widely used pediatric datasets have primarily emphasized disease classification or diagnosis rather than detailed characterization of image confounders, including Pediatric Pneumonia Chest X-ray~\cite{kermany2018identifying}, PediCXR~\cite{pham2023pedicxr}, and VinDr-PCXR~\cite{pham2022vindr}. This limitation is clinically consequential because medical devices are prevalent in PICU or CICU radiographs and can influence model behavior in two ways. First, devices and associated occlusions can directly degrade image-based analysis performance by obscuring relevant anatomy~\cite{liong2022using}. Second, models may learn spurious associations between device presence and target labels, effectively using devices as shortcut features rather than relying on physiologic imaging cues, which can lead to biased or non-generalizable predictions~\cite{rueckel2020impact}. While mask-guided inpainting and related suppression strategies have been well developed in broader computer vision literature~\cite{qin2021image}, their translation to pediatric CXRs has been limited in part due to the scarcity of pediatric datasets with fine-grained, clinically grounded device annotations. \textbf{PediaDeviceSeg-1K is intended to lower this barrier by providing a benchmark resource that supports device-aware modeling, confounder analysis, and downstream tasks beyond device segmentation, thereby facilitating more robust and clinically reliable pediatric CXR research.}

Building on PediaDeviceSeg-1K, we leverage device annotations not only for segmentation benchmarking, but also for enabling device-aware downstream processing within our PCXR-Copilot pipeline. In particular, explicit device masks provide an interpretable representation of non-biologic support hardware and facilitate optional mask-guided device suppression, which can help reduce visual clutter and mitigate device-driven contamination of downstream measurements~\cite{brady2017error, lee2013cognitive,liong2022using, qin2021image}. In parallel, the lung segmentation module delineates the pulmonary fields to focus subsequent analyses on clinically relevant regions, and the edema-related highlight maps provide localized visual evidence that can facilitate rapid inspection and targeted review. In our experiments, the lung and device segmentation modules achieved Dice scores of $92.7 \pm 2.7\%$ and $77.8 \pm 1.8\%$, respectively, supporting their use for pulmonary-region delineation and downstream device-aware processing. Qualitative suppression examples suggest that prominent devices can be substantially reduced while preserving surrounding anatomy in many cases, with predictable failure modes in the presence of thin, low-contrast structures or complex overlap patterns. Together, these results support the feasibility of a device-aware assistance workflow in high-burden pediatric radiographs, while maintaining transparency through intermediate outputs that clinicians can inspect and validate.

Pixel-level reference standards for edema extent are rarely available in routine clinical practice. Accordingly, the goal of our edema-related outputs is not calibrated diagnosis, but rather consistent, inspectable cues that support longitudinal comparison and clinical reasoning. Prior deep learning approaches for pulmonary edema assessment on chest radiographs, developed primarily on adult cohorts, commonly rely on discrete severity labels or ordinal grading schemes~\cite{liao2019semi, schulz2023deep}. While clinically meaningful, such label-supervised predictors typically yield coarse categorical outputs and offer limited transparency regarding the image evidence that drives a given score, which can complicate clinical interpretation and constrain the study of fine-grained within-patient trending. In contrast, we introduce an image-derived pulmonary edema index that is computed directly from the radiograph and within anatomically defined regions, complemented by spatial highlight maps that localize image patterns associated with edema-related signal. This design is intended to offer evidence-grounded interpretability by tying the scalar index to observable intensity characteristics within the lung fields, while providing visual context to support user verification and error checking. Moreover, our approach incorporates device-aware and rib-aware preprocessing to reduce non-parenchymal intensity contamination, and reports both scalar indices and spatial highlight maps to preserve interpretability. In practice, such paired quantitative + visual outputs can provide rapid situational awareness, remove potentially distracting and confounding silhouetting effect from ribs, facilitate targeted image review, and enable hypothesis generation about temporal changes, while remaining amenable to clinician inspection through intermediate masks and overlays produced by the pipeline.

The MIMIC-CXR results show that an edema-related intensity signal remained detectable despite the shift from pediatric development data to an adult external cohort, without additional model training or supervised fitting of the index to edema labels. The need to draw on adult severity-graded data also highlights the shortage of pediatric imaging resources tailored to this task. In this context, PediaDeviceSeg-1K contributes a dedicated pediatric intensive care resource with clinically grounded device annotations, supporting the development and benchmarking of device-aware methods for analyzing radiographs in which support hardware can affect downstream measurements.

Finally, we translated PCXR-Copilot into a real-world clinical workflow by integrating the full pipeline into TWIST at Boston Children’s Hospital. This deployment enables clinicians to access both the final outputs and intermediate results, including lung and device masks, edema highlight maps, scalar indices, and optional device-suppressed views, within the standard image reading environment. By presenting structured, inspectable evidence alongside the original radiographs, the TWIST integration supports practical usability and quality control, and it demonstrated the feasibility of deploying a device-aware pediatric CXR assistance workflow under routine clinical constraints.

Despite these advantages, several limitations still need to be discussed. First, while PediaDeviceSeg-1K provides detailed device annotations across common pediatric ICU hardware, many devices are annotated as polylines to reflect their morphology and to support efficient labeling at scale. For segmentation-based training and evaluation, we convert these annotations into rasterized masks using device-specific width priors that approximate clinically plausible extents. Nevertheless, these derived masks remain an approximation rather than a pixel-level accurate boundary delineation. Second, because routine clinical radiographs are acquired with devices in situ, pixel-aligned device-free reference images are not available, which inherently constrains fully quantitative evaluation of suppression quality and motivates reliance on proxy criteria and qualitative review. Third, the pulmonary edema index is intentionally designed as a descriptive decision-support cue rather than a calibrated physiologic measurement, and its interpretation should be contextualized accordingly. 
Fourth, many radiographs were acquired portably on critically ill ICU patients and were subject to limitations including variable patient positioning, superimposed material, and source-to-image distance; their image quality was therefore likely reduced compared with standardized departmental radiographs. These limitations are inherent to the intended PICU/CICU clinical setting.
Finally, although TWIST integration demonstrates feasibility for real-world deployment, the impact on clinician workflow, time-to-decision, and patient-relevant outcomes remains to be established through prospective, human-in-the-loop evaluation in the intended clinical environment. Future work will therefore focus on prospective clinical studies to quantify usability and impact, and on improving device modeling for broader downstream utility. In particular, expanding robust multi-class device segmentation for rarer categories would enable additional clinically relevant tasks beyond suppression, such as device-specific counting, type-aware position checking, and stratified analyses of device burden when studying pulmonary or postoperative complications.

In conclusion, we present PCXR-Copilot, a device-aware pediatric CXR assistance pipeline that provides structured, inspectable outputs including segmentation masks, optional device suppression, and edema-related quantitative-and-visual cues, and we introduce PediaDeviceSeg-1K to enable reproducible benchmarking and downstream research. Together, these contributions facilitate transparent, device-aware decision support for high-burden pediatric radiographs and provide a foundation for future prospective clinical evaluation and broader adoption.

\section{Methods}

\subsection{Study design and ethics}
This retrospective study developed and technically validated a pediatric chest radiograph (PCXR) assistance pipeline for (i) medical device segmentation and suppression and (ii) edema-related visualization and quantification. The study was conducted at Boston Children’s Hospital (BCH). Institutional review board (IRB-P00037536) approval was obtained for retrospective analysis of de-identified clinical imaging data. All data handling complied with institutional policies for patient privacy and information security.

\subsection{Data source, inclusion criteria, and splits}
The source cohort comprised pediatric patients admitted to the Pediatric Intensive Care Unit (PICU) and Cardiac Surgical Intensive Care Unit (CICU) who underwent routine frontal CXRs as part of clinical care. We included frontal CXRs acquired in routine practice. We excluded non-frontal views (e.g., lateral), non-chest images, and studies with severe corruption or missing metadata that precluded processing. DICOM files were exported via query-retrieve against the vendor neutral archive at BCH, were anonymized, and transferred to the University of Notre Dame.


\subsection{PediaDeviceSeg-1K dataset}
All device annotations were produced using the Computer Vision Annotation Tool (CVAT). Depending on device morphology, annotations were represented either as polylines (for thin, elongated structures) or polygons (for broader device components). The image-level prevalence of each device category in the annotated dataset is summarized in Table~\ref{tab:device_summary}.

Annotations were conducted using a multi-stage quality assurance workflow.
Overall, the annotation and quality-control process spanned \textbf{10 months} from initiation to final expert verification. First, 6 trained annotators performed initial labeling in CVAT, using polylines or polygons depending on device morphology. The initial annotation stage required approximately \textbf{257} person-hours in total, corresponding to an average of \textbf{15.4} minutes per image (range: \textbf{0.4} to \textbf{60.6} minutes). Next, a senior annotator reviewed the full set of annotations and corrected potential errors through targeted re-labeling. This secondary verification stage required approximately \textbf{86} person-hours in total, corresponding to an average of \textbf{5.5} minutes per image (range: $\sim$\textbf{0} to \textbf{25.7} minutes). Finally, an experienced clinician (J.N.K.) with 20 years of clinical experience conducted a final verification pass to ensure label consistency and clinical accuracy.


\subsection{PCXR-Copilot pipeline}
Motivated by the high device burden and visual complexity of pediatric ICU chest radiographs, we developed an end-to-end, device-aware CXR assistance pipeline that combines segmentation, suppression, and edema-related visualization to support clinical interpretation. Given an input frontal pediatric CXR, our proposed pipeline produces a set of structured outputs intended for decision support, comprising lung masks, device masks, device-suppressed images, and edema-related outputs including pulmonary and body-wall edema visualization maps and scalar indices. 

Overall, PCXR-Copilot consists of modular components for lung-field delineation, medical device segmentation, device suppression via mask-guided inpainting, and edema-related outputs. The lung segmentation module defines pulmonary regions of interest, the device segmentation module produces pixel-wise device masks, and the suppression module generates a device-removed radiograph for downstream visualization and quantification. Edema-related cues include a Grad-CAM-based pulmonary edema highlight map and a scalar pulmonary edema index computed within the segmented lung fields, as well as a body-wall edema highlight map and associated width-based summary measures. All outputs are provided to end users through integration with the TWIST clinical platform. In the following subsections, we describe the dataset and annotation workflow, the modeling and training procedures for each module, and the evaluation and deployment methodologies.

\subsubsection{Lung segmentation}
Publicly available pediatric chest radiograph (PCXR) datasets with pixel-wise lung masks remain scarce. To support supervised lung-field segmentation for our PCXR-Copilot pipeline, we manually curated and annotated 300 PCXR images with lung-region masks. We trained and evaluated multiple segmentation backbones using five-fold cross-validation, with U-Net++~\cite{zhou2018unet++} used as the final lung segmentation architecture. To encourage anatomically grounded lung masks and reduce boundary artifacts in low-contrast regions, we combined the pixel-wise cross-entropy loss, $\mathcal{L}_{\mathrm{CE}}$, with an active contour loss, $\mathcal{L}_{\mathrm{AC}}$~\cite{8953484}, and our final loss function is:
\begin{equation}
\mathcal{L}=\mathcal{L}_{\mathrm{CE}}+\mathcal{L}_{\mathrm{AC}}.
\end{equation}


\subsubsection{Medical device segmentation and suppression} 
Device annotations were provided as polylines or polygons and were rasterised to binary masks prior to model training. Polygons were filled directly; polylines were stroked at native resolution with a class-specific width: 10\,px for endotracheal tubes, 8\,px for nasogastric and sump tubes, and 4\,px for all remaining categories.

Device masks were predicted using nnU-Net~\cite{isensee2021nnu} configured as a binary segmentation task (device versus background) and trained with the default planner and hyperparameters on the $512\times512$ label maps described above. During generation of the training label maps, device annotations were dilated with a $5\times5$ elliptical structuring element using a class-specific number of iterations (13 for endotracheal tubes and for chest and Blake drains, 6 for right internal jugular lines, 5 for nasogastric and sump tubes, and 2 for all other categories). Device suppression was performed using a mask-guided inpainting model fine-tuned from LaMa~\cite{suvorov2022resolution}, initialized from the publicly released \texttt{big-lama} checkpoint. The generator is a fast Fourier convolution (FFC) ResNet with 18 residual blocks, 64 base filters, three downsampling stages, an FFC global-to-local ratio of 0.75 within the residual blocks, and a four-channel input formed by concatenating the radiograph with its device mask; the discriminator is a four-layer pix2pixHD patch discriminator. The objective combined an $L_1$ reconstruction term on the unmasked region (weight 10), an $R_1$-regularized adversarial loss (weight 10, gradient-penalty coefficient 0.001), a feature-matching loss (weight 100), and a ResNet-based perceptual loss (weight 30). Both networks were optimized with Adam at learning rates of $5\times10^{-5}$ (generator) and $1\times10^{-4}$ (discriminator), with a per-device batch size of 8 distributed across eight GPUs, gradient clipping at 1.0, and full 32-bit precision. Predicted device masks were supplied directly as inpainting guidance at $512\times512$ resolution, and the reconstructed radiograph was used for all downstream quantification. No additional dilation was applied at inference.


\subsubsection{PE highlighting and PE index}
Pulmonary edema (PE) highlighting was implemented using Grad-CAM computed from a pre-trained ReXNet-150~\cite{han2021rethinking} pulmonary edema classifier. The classifier was trained on a balanced subset of the NIH ChestX-ray14 dataset~\cite{wang2017chestx} comprising 500 pulmonary edema positive and 500 pulmonary edema negative radiographs. Grad-CAM was generated with respect to the edema-positive logit using the last convolutional feature layer as the target layer. The resulting saliency map was normalized to the range $[0,1]$, upsampled to the input resolution, and restricted to the predicted lung fields using the lung mask to produce a lung-field PE highlight map. For visualization, the lung-masked saliency map was overlaid on the input radiograph with fixed opacity.


In addition to the highlight map, we reported a scalar pulmonary edema index as a descriptive quantitative cue. Because pixel-level ground truth for edema extent is not routinely available in clinical practice, we computed the PE index as the mean normalized grayscale intensity within the predicted lung fields (i.e., lung-field ``whiteness''). To mitigate bias from high-intensity non-biologic support devices, this index was computed on the medical-device-suppressed image produced by the inpainting module. Furthermore, because ribs can introduce systematically higher intensities within the lung fields, we reduced rib-driven contributions using an automated rib mask. Specifically, we trained a rib segmentation model based on U-Net++~\cite{zhou2018unet++}, initialized from pretraining on VinDr-RibCXR~\cite{nguyen2021vindr} until convergence, and then fine-tuned using 50 manually annotated images from our cohort. Let $\Omega_L$ denote the predicted lung region and $\Omega_R$ the predicted rib region. We defined the PE index as the mean intensity over the lung region excluding ribs, as:
\begin{equation}
I_{\mathrm{PE}}=\frac{1}{|\Omega_L \setminus \Omega_R|}\sum_{x\in\Omega_L \setminus \Omega_R} I(x),
\end{equation}
where $I(x)\in[0,1]$ denotes the normalized grayscale intensity at pixel $x$. In implementation, this is equivalent to replacing intensities within the predicted rib region $\Omega_R$ by the mean intensity over $\Omega_L \setminus \Omega_R$ prior to averaging, so that rib pixels do not inflate the final score. We emphasize that $I_{\mathrm{PE}}$ is intended for decision support as a relative quantitative indicator rather than a calibrated clinical measurement.

However, the lung-field whiteness index may still vary with acquisition technique (e.g., exposure and tube potential) even when underlying physiology is unchanged. 
To reduce technique-driven variability, when DICOM metadata are available, we further apply a technique normalization using commonly recorded acquisition factors:
\begin{itemize}
    \item \textbf{Exposure} ($E_i$): DICOM (0018, 1152) \texttt{Exposure} in mAs, with fallback to \texttt{ExposureIndex} when \texttt{Exposure} is missing.
    \item \textbf{Tube potential} ($kVp_i$): DICOM (0018, 0060) \texttt{KVP}.
    \item \textbf{Entrance dose} ($D_i$): DICOM \texttt{EntranceDoseInmGy} in mGy.
\end{itemize}
These factors are known to influence detector signal and perceived brightness, with exposure and dose exerting approximately multiplicative effects that are well linearized by a log transform, and $kVp$ capturing energy-dependent penetration over the clinical operating range.

Accordingly, we fit an ordinary least squares (OLS) regression on a calibration set to model technique-associated variation in the per-exam PE index $I_{\mathrm{PE},i}$ for image $i$:
\begin{equation}
I_{\mathrm{PE},i} = \beta_0 + \beta_1 \log(1 + E_i) + \beta_2\, kVp_i + \beta_3 \log(1 + D_i) + \epsilon_i,
\label{eq:tech_fit}
\end{equation}
where $\log(1+\cdot)$ aims to improve numerical stability and reduce the influence of heavy-tailed exposure/dose values. If a subset of technique tags is missing, the corresponding terms are omitted for that exam. When entrance dose is unavailable for a substantial subset, we fall back to the richest identifiable model among $\log(1+E)$, $\log(1+E)+kVp$, and $\log(1+E)+kVp+\log(1+D)$, selected automatically based on data availability.

After estimating $\hat{\beta}$ on the calibration set, we compute a technique-adjusted PE index by subtracting the predicted technique component and adding back the calibration-set mean to preserve the original scale, as:
\begin{equation}
\tilde{I}_{\mathrm{PE},i}
= I_{\mathrm{PE},i} - \Big[\hat{\beta}_0 + \hat{\beta}_1 \log(1 + E_i) + \hat{\beta}_2\, kVp_i + \hat{\beta}_3 \log(1 + D_i)\Big] + \overline{I}_{\mathrm{PE},\mathrm{cal}},
\label{eq:tech_norm}
\end{equation}
where $\overline{I}_{\mathrm{PE},\mathrm{cal}}$ denotes the mean of $I_{\mathrm{PE}}$ on the calibration set. Adding $\overline{I}_{\mathrm{PE},\mathrm{cal}}$ preserves the original scale of the index for interpretability across figures and cohorts, whereas setting the additive constant to $0$ yields pure residuals. When $E_i$ is not available for a given exam, normalization is not applied and we report the raw $I_{\mathrm{PE},i}$. To avoid leakage, $\hat{\beta}$ is estimated using training data only (or via cross-fitting) and then held fixed for validation, test, and deployment. Refitting is recommended when acquisition protocols and/or equipment change.

\subsubsection{External evaluation in MIMIC-CXR}

Publicly available pediatric chest radiograph datasets provide limited resources for pulmonary edema severity evaluation. For example, PediCXR contains 9,125 radiographs but includes only one image labeled as edema in its training set and none in its test set; its annotations describe finding presence and localization and do not include ordinal edema-severity grades~\cite{pham2023pedicxr}. We therefore selected adult MIMIC-CXR radiographs with publicly released edema severity grades to evaluate transfer beyond the pediatric development cohort~\cite{johnson2019mimic,liao2021edemagrades,horng2021deep}. This evaluation used our existing pediatric lung segmentation model with fixed weights and an intensity index that required no supervised fitting to edema labels; no model was retrained or fine-tuned on MIMIC-CXR.

We used images from MIMIC-CXR-JPG linked to the severity annotations curated within a congestive heart failure cohort. Restricting images to AP or PA views recorded in the metadata yielded 653 radiographs from 270 patients: 133 with image-level radiologist-consensus grades and 520 with study-level grades assigned by experts from radiology reports. Image-level labels were linked by image identifier, and report-derived labels by patient and study identifiers. Severity was coded as 0 (none), 1 (vascular congestion), 2 (interstitial edema), and 3 (alveolar edema).

Grayscale images and predicted masks were resampled to $640\times640$ for measurement. Images were excluded if the combined lung mask contained fewer than 1,000 pixels or either lung contained fewer than 200 pixels, leaving 648 radiographs from 268 patients, including 131 with consensus grades and 517 with report-derived grades. For this external analysis, the index was the equally weighted mean of the two per-lung mean grayscale intensities on the 0--255 scale, calculated without device suppression, rib exclusion, or acquisition-parameter normalization.

Association with the four-level severity grade was assessed using Spearman correlation. Discrimination between grades 0 and 3 was assessed using ROC AUC, with grade 3 as the positive class. Percentile 95\% confidence intervals were obtained from 2,000 bootstrap resamples of patients with replacement, retaining all eligible radiographs belonging to each sampled patient (random seed 0).

\subsubsection{Body-wall edema highlighting}
Body-wall edema (BWE) highlighting was implemented as a supervised segmentation task using U-Net++ with an EfficientNet-B0~\cite{tan2019efficientnet} encoder and single-channel input. To obtain pixel-level supervision, we manually annotated 60 frontal pediatric CXRs with bilateral body-wall edema regions. A target region was defined as the lateral soft-tissue band in the axillary and chest-wall area, bounded laterally by the external body silhouette and medially by the lateral rib-cage margin, extending inferiorly to roughly the level of the costophrenic angles/hemidiaphragms, while excluding abdominal soft tissues below the diaphragmatic contour, as illustrated in Fig.~\ref{fig:bwe}a. All annotations were stored as binary masks and used to train the segmentation model with five-fold cross-validation; unless otherwise noted, probability maps were binarized at a threshold of 0.5 for downstream visualization and quantification.

In rare cases, as shown in Fig.~\ref{fig:bwe}b, the raw model prediction contained more than two disconnected regions due to small spurious fragments or partial duplications. To enforce a clinically consistent bilateral output, we applied a post-processing step that (i) removes small connected components below a minimum area fraction, and (ii) retains at most two components, prioritizing regions with the largest vertical extent. 

We quantified BWE severity by computing a per-side chest-wall width in physical units. For each side, we computed row-wise mask widths as the horizontal span of the binary mask at each image row, and summarized this distribution using a robust statistic, namely, the mean over the interquartile range from the 25th to the 75th percentile. Pixel widths were converted to millimeters using DICOM pixel spacing, with an additional correction for image resizing when needed. The final BWE index for an image was defined as the mean of the left-side and right-side widths in millimeters. For reporting consistency with patient anatomy, the left and right values were mapped to patient-left and patient-right, noting that image-left corresponds to patient-right under standard radiographic display conventions.

\subsection{Evaluation metrics and statistical analysis}
All supervised segmentation tasks, including the lung, rib, medical device, and body-wall edema segmentation, were evaluated using the Dice coefficient. Results were reported with five-fold cross-validation as mean $\pm$ SD across the held-out folds. Unless otherwise noted, probability maps were binarized at a threshold of $0.5$ prior to scoring.

Quantitative evaluation of device suppression is challenging because radiographs were acquired with devices in situ, and a pixel-aligned device-free reference of the same acquisition is not available. We therefore used the following proxy criteria that do not require device-free ground truth: (i) \emph{removal effectiveness}, measured as the residual device mask area obtained by re-running the device segmenter on the suppressed image within the original device region, and (ii) \emph{anatomical preservation}, measured as the mean absolute intensity change outside the device mask. We additionally report representative qualitative examples and failure cases to characterize typical artifacts and operational boundaries.


For scalar edema-related indices (e.g., pulmonary edema index and body-wall edema index), we report descriptive statistics (median [IQR] and mean $\pm$ SD). To obtain an exploratory clinical reference for directional within-patient change, we selected the first and last frontal radiographs from each patient's longest admission, defined by chaining consecutive examinations separated by no more than 7 days, yielding 97 image pairs. Pair selection was based solely on this temporal criterion and was independent of the PE index. Each pair was displayed side by side without per-image contrast adjustment, with left--right position randomized independently of temporal order. A pediatric cardiac intensivist (J.N.K.), blinded to the PE index, judged which image showed greater pulmonary edema or whether the two appeared approximately the same. For pairs receiving a directional judgment, ROC AUC was computed using the signed between-image difference in the PE index. To assess device-related sensitivity of the pulmonary edema index, we characterized its association with device--lung overlap derived from the predicted device and lung masks using Spearman correlation, with devices either retained or suppressed under otherwise identical processing conditions.

All experiments and analyses were performed in Python V3.10.18, and all reported results were generated using fixed random seeds and deterministic data splits to support reproducibility.

\subsection{Clinical deployment in TWIST}
Following ingestion of each chest radiograph, inference is executed in a containerized service with fixed model versions and deterministic preprocessing. Outputs are returned as image overlays and derived quantitative summaries for visualization alongside the original radiograph.



\section{Data availability}
PediaDeviceSeg-1K, together with additional datasets used to develop components of the PCXR-Copilot pipeline (including 300 pediatric CXRs with lung masks, 50 pediatric CXRs with rib masks, and 60 pediatric CXRs with body-wall edema masks), will be released in a public repository upon publication. During peer review, these data are available to the Editors and reviewers upon request. Any applicable institutional restrictions and required data-use terms will be described in the repository record.

The MIMIC-CXR-JPG images are available through PhysioNet (\url{https://physionet.org/content/mimic-cxr-jpg/}), and the pulmonary edema severity annotations are available at \url{https://physionet.org/content/mimic-cxr-pe-severity/1.0.1/}. Access is governed by the respective PhysioNet credentialing and data-use requirements.
\section{Code availability}
The PCXR-Copilot pipeline, together with trained model weights used in this study (including segmentation and suppression models), will be released in a public repository upon publication. During peer review, the code and model weights are available to the Editors and reviewers upon request.

\section{Author contributions}
X.L. conceived the study, developed the computational methodology and PCXR-Copilot pipeline, performed the experiments and statistical analyses, coordinated the annotation process, participated in data annotation and validation, prepared the figures, and drafted the manuscript. A.G. contributed to study design, clinical data collection and curation, clinical interpretation, data annotation and label review, clinical guidance, and manuscript revision. P.Z. led the integration and deployment of PCXR-Copilot within the TWIST clinical platform. M.J. developed the medical device suppression component and contributed to device segmentation experiments. S.H. coordinated the annotation process and participated in data annotation. G.X. contributed to methodological development and technical guidance. D.Z.C. supervised methodological development, contributed to study design and interpretation of the results, and critically revised the manuscript. Y.S. supervised the computational aspects of the study, contributed to study conception and methodology, and critically revised the manuscript. J.N.K. supervised the clinical aspects of the study and contributed to clinical data collection and curation, clinical interpretation, annotation and label review, clinical guidance, and manuscript revision. X.L. and A.G. contributed equally to this work. Y.S. and J.N.K. jointly supervised this work. All authors reviewed and approved the final manuscript.

\section{Competing interests}
The authors declare no competing interests.

\bibliography{sn-bibliography}

\end{document}